\documentclass{article}
\usepackage{ijcai22}
\usepackage{algorithm}
\usepackage{algorithmic}
\usepackage{times}

\usepackage{soul}
\usepackage{url}
\usepackage[hidelinks]{hyperref}
\usepackage[utf8]{inputenc}
\usepackage[small]{caption}
\usepackage{graphicx}
\usepackage{amsmath, bm}
\usepackage{svg}
\usepackage{booktabs}
\usepackage{subcaption}
\usepackage{amsfonts}       
\usepackage{multirow}
\usepackage{multicol}
\usepackage{makecell}
\usepackage{mathptmx}
\usepackage{boldline}
\usepackage{graphics}
\usepackage{siunitx}
\usepackage{rotating}
\usepackage{etoolbox}
\usepackage{float}
\usepackage{placeins}
\usepackage{longtable}

\DeclareCaptionType{equ}[][]

\title{Conditional Variational Autoencoder with Balanced Pre-training for Generative Adversarial Networks }

\author{
Yuchong Yao$^1$\and
Xiaohui Wangr$^1$\and
Yuanbang Ma$^1$\and
Han Fang$^1$\and
Jiaying Wei$^1$ \and
Liyuan Chen$^1$\and
Ali Anaissi $^1$\And
Ali Braytee $^{1,2}$\\
\affiliations
$^1$School of Computer Science, The University of Sydney\\
$^2$School of Computer Science, University of Technology Sydney\\
}

\begin{document}

\maketitle

\begin{abstract}
Class imbalance occurs in many real-world applications, including image classification, where the number of images in each class differs significantly. With imbalanced data, the generative adversarial networks (GANs) leans to majority class samples. The two recent methods, Balancing GAN (BAGAN) and improved BAGAN (BAGAN-GP), are proposed as an augmentation tool to handle this problem and restore the balance to the data. The former pre-trains the autoencoder weights in an unsupervised manner. However, it is unstable when the images from different categories have similar features. The latter is improved based on BAGAN by facilitating supervised autoencoder training, but the pre-training is biased towards the majority classes.
In this work, we propose a novel \textit{Conditional Variational Autoencoder with Balanced Pre-training for Generative Adversarial Networks} (CAPGAN)\footnote{We will open source the code upon acceptance} as an augmentation tool to generate realistic synthetic images. In particular, we utilize a conditional convolutional variational autoencoder with supervised and balanced pre-training for the GAN initialization and training with gradient penalty.  Our proposed method presents a superior performance of other state-of-the-art methods on the highly imbalanced version of MNIST, Fashion-MNIST, CIFAR-10, and two medical imaging datasets. Our method can synthesize high-quality minority samples in terms of Fréchet inception distance, structural similarity index measure and perceptual quality. 
\end{abstract}

\section{Introduction}

Computer vision contains many supervised learning problems, including image classification, image segmentation, and others\cite{he2016deep}. Modern image classifiers are generally deep learning models which need balanced imaging datasets such as MNIST\cite{lecun2010mnist}, Fashion-MNIST\cite{xiao2017fashionmnist}, CIFAR-10\cite{krizhevsky2009learning}, ImageNet\cite{deng2009imagenet}, and others. However, class distribution in real-world datasets is often skewed, especially in the medical imaging domain, i.e. there are more normal images than cancerous images. The performance of the image classification models based on deep learning degrade significantly in the presence of class imbalance because these models will be biased towards the majority class samples and ignore the minority ones~\cite{braytee2019correlated}.

Interestingly, the literature shows that augmenting the minority classes with sample generation such as Generative Adversarial Networks (GANs) in image generation is a promising approach to deal with class imbalance~\cite{rezaei2020generative}. Briefly, GANs consist of a generator and a discriminator that adopt an adversarial training schema to allow the generator and discriminator to compete. GANs can generate synthetic minority samples to help restore balance to the data. However, GANs-based approaches have several limitations, including mode collapse, sub-optimal initialization, and training instability, which lead to unstable results. Further, bias can occur towards the majority class. Recent works combine GANs with other models, such as autoencoder, to borrow the reconstruction ability to enhance the initialization and training of the GANs-based models for generating minority samples. A recent powerful methods BAGAN \cite{mariani2018bagan}, and BAGAN-GP \cite{huang2021enhanced} are examples that have shown promising results to handle class imbalance on various benchmarks and datasets. Nevertheless, they still suffer from the following limitations. The pre-training in BAGAN-GP is created on imbalanced data which lead to be biased towards the majority classes. Further, BAGAN-GP used naive autoencoder model and objective function to obtain the pre-trained weights, which can be further optimized by more advanced architectures and objectives. Moreover, BAGAN and BAGAN-GP are lack of comprehensive evaluations, where they only evaluated the models under datasets with a small imbalance rate and they haven't evaluated on highly and extreme imbalance rates. To this end, we propose a new framework, namely, \textit{Conditional Variational Autoencoder with Balanced Pre-training for Generative Adversarial Networks} (CAPGAN). The general objective of our framework is to synthesize high-quality samples for the majority and minority classes to overcome the class imbalance problem. 

The major contributions of CAPGAN can be summarized as follows: (1) we facilitate a balanced pre-training stage to GAN components; (2) we utilize conditional variational autoencoder model in the pre-training stage for GAN initialization; (3) we propose a novel sophisticated objective function that encourages the model to capture the true distribution of the samples and generate high-quality samples; (4) we integrate the proposed balanced pre-training and the new objective function simultaneously to initialize and train the corresponding GAN components to enhance the training stability and generate more realistic and diverse samples.

\section{Related Work}

\textbf{Generative Models on Class Imbalance.}
Autoencoder and generative adversarial networks (GANs) are two representative generative models proposed to handle class imbalance in imaging applications. 
Several studies suggest that acquiring more samples (especially the minority classes) to restore data balance is the most effective way to address the class imbalance. Few studies use autoencoder variants to handle class imbalance. For example, Taghanaki et al. (2020) state that variational autoencoder (VAE) can improve the performance on imbalanced data~\cite{taghanaki2020jigsaw}. However, Li et al. (2018) find that the samples generated by VAE are not as diverse as the samples from GANs~\cite{li2018imbalanced}. Hence, several studies finds GANs variants are powerful to handle the imbalanced data. DCGAN~\cite{shoohi2020dcgan} is proposed to synthesize samples for the minority classes. It leads to impressive results on various tasks (e.g. plant disease). WGAN is also widely used for data augmentation \cite{bhatia2019using} and minority oversampling for CT images\cite{wang2019wgan}. CycleGAN applies image-to-image translation on the imbalanced data, which attempts to generate minority samples based on majority samples\cite{zhu2017unpaired}. Many existing studies attempt to facilitate semi-supervised GANs (or conditional GANs) and unsupervised GANs together. For example, \cite{balasubramanian2020analysis} uses an unconditional GAN for diabetes image oversampling. Further, another study proposes a conditional GAN (CovidGAN)~\cite{waheed2020covidgan} to augment the minority Covid-19 CXR images. Although the generative models achieve impressive results for addressing the class imbalance, they suffer from mode collapse, training instability, and unstable results. Further, some studies argued that the generated minority samples would bias towards the majority classes and degrade the original performance for majority samples \cite{sampath2021survey}. 

\textbf{GAN-Autoencoder-based augmentation}
To overcome the limitation of GANs augmentation methods to handle the class imbalance, BAGAN combines the power of autoencoder and GANs~\cite{mariani2018bagan}. It integrates an autoencoder with GANs to gain better reconstruction ability and produces a stable starting point for training. Particularly, it initializes the GAN model by integrating the pre-trained autoencoder (given that the GANs and the autoencoder have the same network architectures). BAGAN can only uses an unsupervised autoencoder where it does not use the label information during the pre-training. However, label information is critical for imposing class conditioning on pre-trained weights. Recently, an improved version of BAGAN (i.e. BAGAN-GP)~\cite{huang2021enhanced} states that BAGAN does not perform well on medical data and the results are not stable. BAGAN-GP introduces a supervised autoencoder, which utilizes the label information during pre-training. Besides, BAGAN-GP applies conditional GANs in its structure to improve the class-specific generative performance. The results show that BAGAN-GP is superior over BAGAN-GP on various datasets (MNIST, Fashion-MNIST, and CIFAR-10). Furthermore, BAGAN-GP is proved to be more effective than BAGAN on the medical imaging data. However, BAGAN-GP facilitates simple network structures and basic objective functions in its autoencoder. Also, although the pre-training takes class information into account, the pre-training is biased towards majority classes, leading to sub-optimal solutions.

\section{Method}



\begin{figure*}[t]
    \centering
    \begin{subfigure}[t]{0.45\linewidth}
        \centering
    \includegraphics[width=\linewidth]{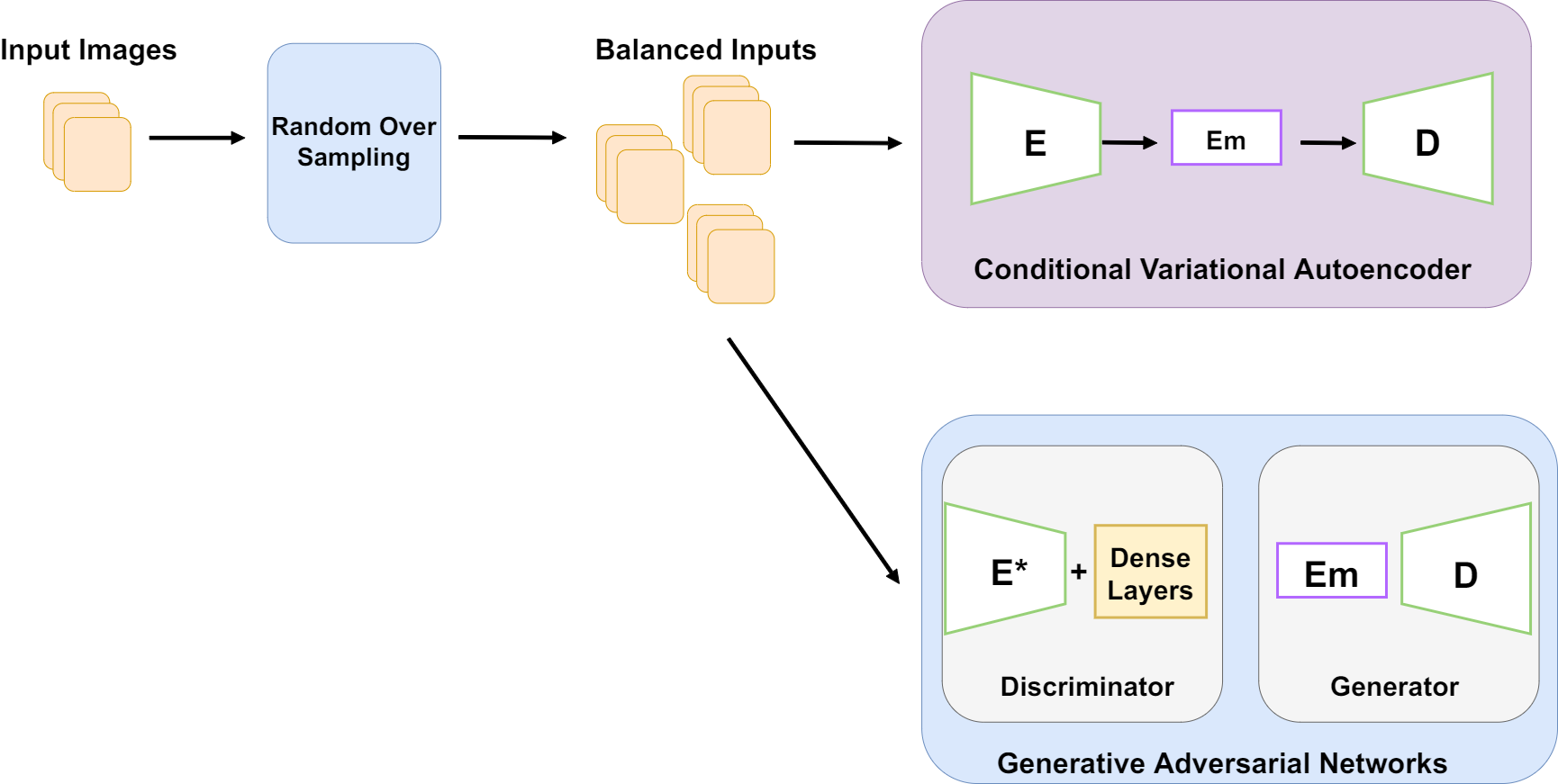}
    \caption{Supervised Initialization and Training}
    \end{subfigure}
    \hfill
    \begin{subfigure}[t]{0.5\linewidth}
        \centering
    \includegraphics[width=\linewidth]{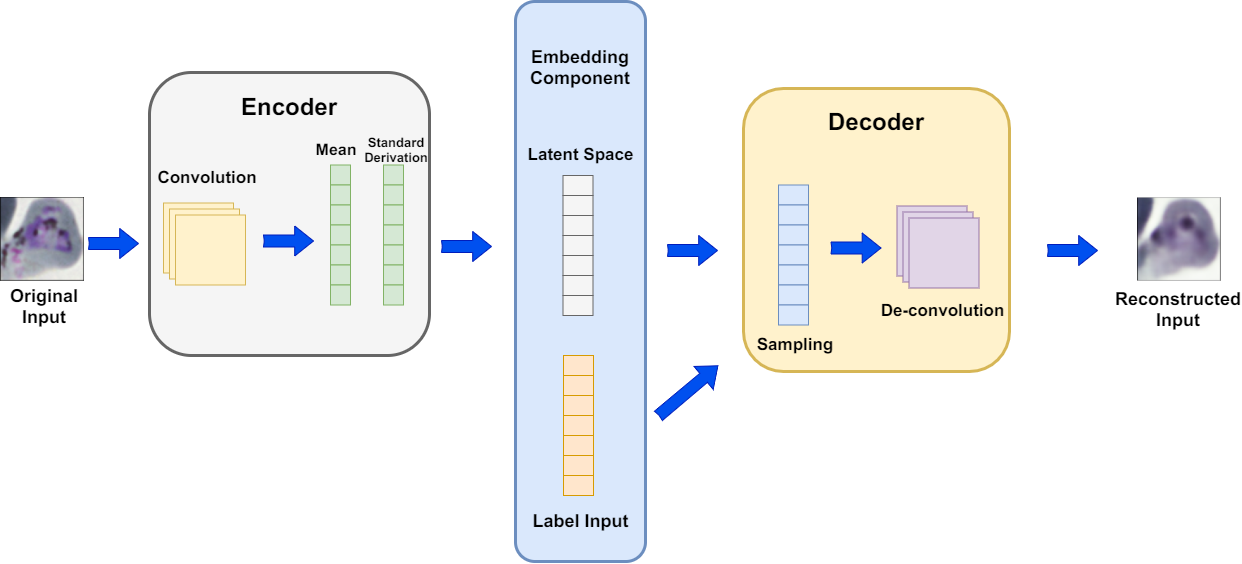}
    \caption{Improved Conditional Variational Autoencoder in CAPGAN}
    \label{fig:CVAE}
    \end{subfigure}
    \hfill

    \subcaptionbox{Pre-training strategies (Left: Oversampling. Middle: Two-phase Pre-training. Right: Ensemble Pre-training.)\label{fig:Pre-training}}%
    {
    \begin{subfigure}[t]{0.3\linewidth}
     \centering
     \includegraphics[width=\linewidth]{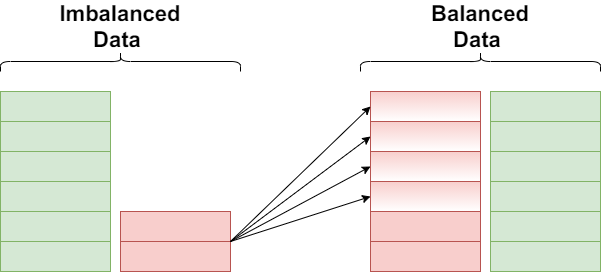}
    
     \end{subfigure}%
     \hfill
     \begin{subfigure}[t]{0.3\linewidth}
         \centering
         \includegraphics[width=\linewidth]{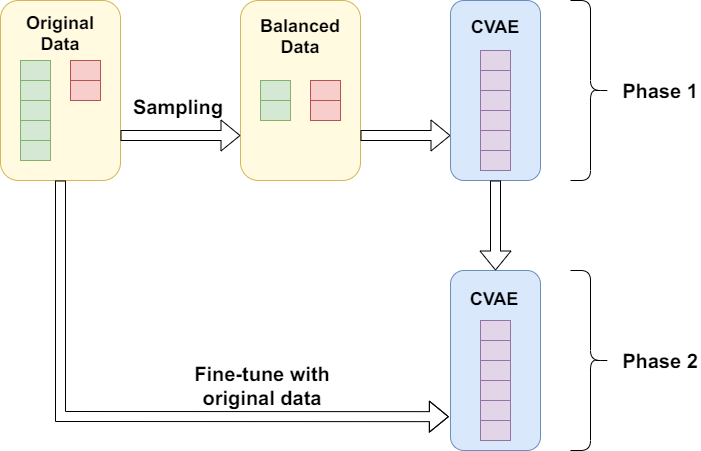}
     \end{subfigure}
     \hfill
     \begin{subfigure}[t]{0.3\linewidth}
         \centering
         \includegraphics[width=\linewidth]{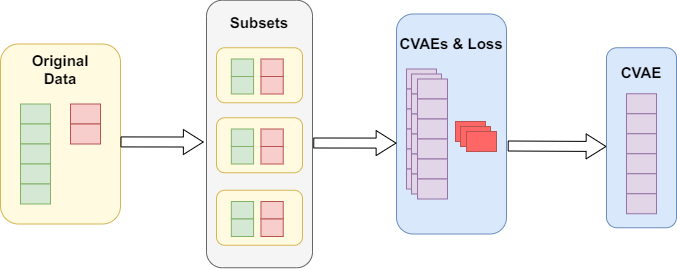}
        
     \end{subfigure}
     
     }
     \caption{Our proposed CAPGAN framework}
    \label{fig:CAPGAN_Framework}
\end{figure*}

\subsection{Supervised Initialization and Training}

We initialize the discriminator and the generator in GAN with the weights of a pre-trained conditional variational autoencoder (CVAE). When the GANs are trained under the adversarial settings, the generator can produce class-specific samples with better representative ability given by the pre-training, especially for the minority classes. Also, the discriminator can identify whether the image belongs to one of the classes or is a fake sample. In this step, the discriminator and generator are updated in GAN by following the min-max adversarial settings\cite{goodfellow2014generative} in Equation~\ref{equ:Adversatial Training}

\begin{equ}[!ht]

\begin{equation}
    \begin{split}
        \min_{G}\max_{D}V(D,G) &= \mathbb{E}_{x\sim p_{data}(x)}[logD(x)]\\  
  &+\mathbb{E}_{z\sim p_z(z)}[log(1-D(G(z)))]  
    \end{split}
\label{equ:Adversatial Training}
\end{equation}

\end{equ}

where $z$ denotes noise samples, and $ x $ is from data generating distribution.
After we develop CVAE in the initialization step, the weights of CVAE's components are transferred to GAN's components. The generator and the decoder (with the embedding component) are designed to have the same network structures and topologies to allow the weights to be transferred from the pre-trained decoder to the generator. The discriminator is initialized to have the same network structures and topologies in the first few layers as the encoder, followed by dense layers that match the dimension in the final output. The weights of the final dense layers are randomly initialized. The initialization of the generator and discriminator in GAN is illustrated in Figure \ref{fig:CAPGAN_Framework}(a).

Training the CAPGAN follows the standard adversarial settings that the discriminator and the generator compete with each other. The loss function of the discriminator and the generator is inspired from DRAGAN \cite{kodali2017convergence}. The discriminator consists of three losses for fake images, real images, and wrong labels, while the generator has only a fake image loss. Moreover, we impose the gradient penalty term\cite{arjovsky2017wasserstein} in the discriminator loss, which aims to help the convergence of the discriminator as shown in Equation~\ref{equ:gradient penalty}.

\begin{equ}[!ht]
\begin{equation}
GP = \lambda \mathbb{E}_{\hat{x} \sim \hat{X}}[(\left \| \nabla_{\hat{x}}  D(\hat{x})\right \|_2 -1)^2]
\label{equ:gradient penalty}
\end{equation}

\end{equ}
where  $\hat{x} = \alpha x_r+(1-\alpha)x_{noise},\alpha \sim \mathrm {U}(0,1) $, $x_r$ is the real input image, $\alpha$ is a normally distributed random number with values in uniform distribution $ \mathrm {U}(0,1)$, $\lambda$ denotes gradient penalty weight and $\left \| \nabla_{x}  D(x)\right \|_2$ is the norm of gradients. Both the discriminator and the generator attempt to minimize their losses to obtain better performance. 

\subsection{Improved Conditional Variational Autoencoder in CAPGAN}
The simple autoencoder utilized by BAGAN-GP is replaced with a more powerful Conditional Convolutional Variational Autoencoder in CAPGAN. 
Variational autoencoder has an advanced architecture compared to the autoencoder. Autoencoder may create some samples in the latent space with no valid meaning or hard to interpret after decoding, leading to a poor generative performance for the decoder in creating new samples from the latent space. However, variational autoencoder overcomes latent space irregularity by encoding the inputs into mean and standard derivation (i.e. learns a distribution over the latent space). Hence, the latent space is continuous, regularized, and enables easier sampling and interpolation. Reparametrization is applied to integrate the learned mean and standard derivation from the encoder in CVAE as described in equation \ref{equ:argmin}, \ref{equ:musigma}, \ref{equ:labeled_z}. Convolutional layers and transposed convolutional layers are utilized in the encoder and the decoder of CVAE. There is an embedding component (a shallow sub-network) in our proposed CVAE that takes the class labels of the input images and encodes them into class-specific information (i.e. same size as the latent space). The outputs from the encoder and the embedding component are fed into the decoder together as the input, where the latent output and the class-specific information are multiplied Equation \ref{equ: embedded_output}
 



 
 


\begin{equ}[!ht]

  \begin{equation}
      \begin{split}
          (\mathfrak g^*,\mathfrak h^*) &= \arg \min\  KL(\mathfrak q_x(y),p(\mathfrak y\mid x )) \\
        & = \arg \min (\mathbb{E}_{y\sim \mathfrak q_x}(log\ \mathfrak q_x(y))-\mathbb{E}_{y\sim \mathfrak q_x}(log\frac{\mathfrak p(x\mid y)\mathfrak p(y)}{\mathfrak p(x)}))  \\
        & = \arg \max (\mathbb{E}_{y\sim \mathfrak q_x}(log\ \mathfrak p(x\mid y))-KL(\mathfrak q_x(y),\mathfrak p(y) )  
      \end{split}
      \label{equ:argmin}
  \end{equation}
  
    \begin{equation}
        \begin{split}
           &\mu _x = \mathfrak g(x)=\mathfrak g_2(\mathfrak g_1(x)),\\
 &\sigma _x= \mathfrak h(x) =\mathfrak h_2(\mathfrak h_1(x))\\
 &\ \mathfrak g_1(x)=\mathfrak h_1(x)
        \end{split}
        \label{equ:musigma}
    \end{equation}
\begin{equation}
    z=\mu+ \exp(0.5\times \sigma)\times \alpha,\ \ \alpha \sim \mathrm {U}(0,1)
    \label{equ:labeled_z}
\end{equation}
\begin{equation}
\mathrm{O} =z \odot e(z,\mathrm{y}) 
\label{equ: embedded_output}
\end{equation}
\label{equ:Reparametrisation Trick}
\end{equ}
where $z$ is the reparameterized variable in the latent dimension with a mean $\mu$ and standard derivation $\sigma$, $\alpha$ is a normally distributed random number with values in uniform distribution $ \mathrm {U}(0,1)$, $e$ and $\mathrm{y}$ denotes the embedding component and class label respectively, $\mathbf{O}$ is the embedded output.
In this way, we manage to pass the class information into the CVAE and train it in a supervised fashion. This is critical for the GAN initialization as the training of the GAN is supervised. Transferring weights from an unsupervised CVAE could mislead the GAN and result in sub-optimal solutions. Therefore, it is essential to facilitate the embedding component to make CVAE training conditioned on the class labels. The illustration of the proposed CVAE architecture component in CAPGAN is shown in Figure \ref{fig:CVAE}.

\subsection{Improved Objective Function}
The BAGAN-GP method applies L2 minimization to train autoencoder, but this may lead to two drawbacks: firstly, mean absolute error (MAE) only enforces pixel-to-pixel similarity, which fails to capture the class-wise distributions of the input samples. Secondly, MAE is not suitable for training more advanced and sophisticated CVAE \cite{kingma2013auto}.
The new objective function of our proposed CVAE model is composed of three components: (1) the Kullback–Leibler (KL) divergence; (2) the cross-entropy loss; (3) and the mean squared error. KL-divergence measures the difference between two probability distributions which is considered critical for training the CVAE because it encourages the model to learn a distribution in the latent space. By minimizing the KL-divergence, the learned mean and standard derivation for the target distribution of latent space are optimized, which allows the decoder to sample and generate better results. Further, the KL-divergence is denoted as the latent loss in the objective function. The remaining two components are related to the reconstruction loss. Cross entropy loss is more suitable for Bernoulli distribution as it expresses the negative Bernoulli log-likelihood, while the mean squared error assumes a Gaussian distribution. Incorporating these two components is significant to optimize the model’s performance on more complex distributions. By minimising both cross-entropy loss and mean squared loss, the CVAE can learn better reconstruction ability on more sophisticated distributions and gain better generative performance. The objective function is presented in Equation~\ref{eq:objective Function} as follows

\begin{equ}[!ht]
\begin{equation}
     Objective = D_{KL}( p\parallel q) + H(p,q) + MSE
\label{eq:objective Function}     
\end{equation}

\begin{equation}
D_{KL}( p\parallel q  ) = \sum_{x}^{} p ( x )\ln_{}{\frac{p(x)}{q(x)} }\label{equ:KL}
\end{equation}  
\begin{equation}
    H(p,q) = -\sum_{x}^{}p(x)\log_{}{q(x)}
\end{equation}
\begin{equation}
    MSE = \frac{1}{m}\sum_{i=1}^{m}(x_i - \hat{x}_i)^2
\end{equation}

\label{equ:objective Function}
\end{equ}

\subsection{Random Oversampling Pre-Training Strategy}
The discriminator and the generator in the GAN require a good initialization point to synthesize balanced samples towards all classes. The original autoencoder pre-training strategy in the BAGAN-GP method is imbalanced towards the majority classes, which introduces burdens for the GAN training. In CAPGAN, we redesigned the pre-training strategy to allow the pre-trained weights to be balanced for the GAN initialization. Three different pre-training strategies are explored as show in Figure \ref{fig:Pre-training}. Two strategies that are implemented but not adopted called two-phase pre-training and ensemble pre-training. The two-phase pre-training resembles the ideas from two-phase learning, where the CVAE is first trained on balanced data, then the CVAE is fine-tuned on the original imbalanced dataset. However, this approach suffers from overfitting and high cost on tuning. The ensemble strategy attempts to fit multiple CVAEs with different subsets of the majority classes and combines them with the entire minority samples. The final weights would be a weighted average of all weights from the CVAEs according to their training loss. This method has drawbacks such as computationally expensive and overfitting and it may be infeasible in practice.

In CAPGAN, we adopt random oversampling (ROS) for pre-training strategy. ROS has been proved to be effective in many applications for addressing class imbalance. The simplicity and compatibility of the method make it a popular choice for many class imbalance applications. The imbalanced data is randomly oversampled to make samples in each class are balanced before they are fed into the CVAE. Although there is a potential risk for overfitting the minority samples as they are replicated multiple times, the ROS pre-training shows an improved performance with minor computational costs. 
The reason is due to transferring the weights from the CVAE to the GAN only during the initialization step.
These weights from the ROS pre-training strategy manage to produce a balanced and good enough starting for the GAN to achieve great results. Furthermore, ROS pre-training can be easily scaled to large and complex datasets due to its simplicity and computational efficiency.

\begin{table*}[t]
\centering

\resizebox{\linewidth}{!}{%
\begin{tabular}{rlrrrrrrrrrrrr}
\toprule
                                                                              &                                                           & \multicolumn{4}{c}{\textbf{MNIST}}                               & \multicolumn{4}{c}{\textbf{Fashion-MNIST}}                       & \multicolumn{4}{c}{\textbf{CIFAR-10}}                             \\ \cmidrule{2-14} 
                                                                              &                                                           & \multicolumn{2}{c}{avg(Minority)} & \multicolumn{2}{c}{Majority} & \multicolumn{2}{c}{avg(Minority)} & \multicolumn{2}{c}{Majority} & \multicolumn{2}{c}{avg(Minority)} & \multicolumn{2}{c}{Majority} \\
 &  & FID & SSIM & FID & SSIM & FID & SSIM & FID & SSIM & FID & SSIM & FID & SSIM \\
\begin{tabular}[c]{@{}r@{}}Imbalance\\ Rate\end{tabular} & Model &  &  &  &  &  &  &  &  &  &  &  &  \\ \midrule

\multirow{3}{*}{5} & DCGAN & 251.33 & \num{2.57E-01} & 207.93 & $\mathbf{2.96\times10^{-1}}$ & 379.96 & \num{2.60E-01} & 316.61 & \num{2.79E-01} & 495.27 & \num{5.67E-02} & 335.43 & $\mathbf{9.30\times10^{-2}}$ \\
 & BAGAN-GP & 176.15 & \num{2.56E-01} & 157.39 & \num{2.86E-01} & 267.05 & $\mathbf{2.67\times10^{-1}}$ & 240.21 & \num{2.82E-01} & 463.85 & \num{5.78E-02} & 364.04 & \num{8.94E-02} \\
 & CAP-GAN & \textbf{165.59} & $\mathbf{2.69\times10^{-1}}$ & \textbf{157.16} & \num{2.79E-01} & \textbf{264.65} & $\mathbf{2.67\times10^{-1}}$ & \textbf{232.47} & $\mathbf{2.88\times10^{-1}}$ & \textbf{366.06} & $\mathbf{6.00\times10^{-2}}$ & \textbf{288.58} & \num{8.00E-02} \\ \cmidrule{2-14} 
 
\multirow{3}{*}{10} & DCGAN & 213.39 & \num{2.58E-01} & 528.09 & \num{3.36E-01} & 472.96 & \num{2.53E-01} & 293.46 & \num{2.96E-01} & 495.27 & \num{5.67E-02} & 335.43 & $\mathbf{9.30\times10^{-2}}$ \\
 & BAGAN-GP & 187.44 & \num{2.62E-01} & 165.08 & \num{2.70E-01} & 328.19 & \num{2.66E-01} & 254.57 & $\mathbf{3.03\times10^{-1}}$ & 486.36 & \num{5.76E-02} & 354.85 & \num{9.06E-02} \\
 & CAP-GAN & \textbf{177.22} & $\mathbf{2.64\times10^{-1}}$ & \textbf{160.12} & $\mathbf{2.90\times10^{-1}}$ & \textbf{271.89} & $\mathbf{2.75\times10^{-1}}$ & \textbf{241.99} & \num{2.94E-01} & \textbf{399.1}8 & $\mathbf{6.00\times10^{-2}}$ & \textbf{291.54} & \num{8.00E-02} \\ \cmidrule{2-14}

\multirow{3}{*}{20} & DCGAN & 278.80 & \num{2.68E-01} & 210.22 & \num{2.85E-01} & 509.82 & \num{2.67E-01} & 468.58 & \num{2.25E-01} & 659.34 & \num{5.48E-02} & 460.03 & $\mathbf{8.69\times10^{-2}}$ \\
 & BAGAN-GP & 193.40 & \num{2.66E-01} & 174.21 & \num{2.78E-01} & 325.61 & $\mathbf{2.69\times10^{-1}}$ & 288.96 & \num{2.87E-01} & 529.59 & \num{5.48E-02} & 480.33 & \num{7.40E-02} \\
 & CAP-GAN & \textbf{173.59} & $\mathbf{2.70\times10^{-1}}$ & \textbf{149.04} & $\mathbf{2.86\times10^{-1}}$ & \textbf{260.61} & \num{2.64E-01} & \textbf{230.71} & $\mathbf{2.98\times10^{-1}}$ & \textbf{368.46} & $\mathbf{6.00\times10^{-2}}$ & \textbf{271.10} & \num{8.00E-02} \\ \cmidrule{2-14}

\multirow{3}{*}{50} & DCGAN & 479.63 & \num{2.55E-01} & 722.07 & \num{3.13E-01} & 540.12 & $\mathbf{2.74\times10^{-1}}$ & 514.94 & \num{2.65E-01} & 742.54 & \num{5.71E-02} & 453.33 & $\mathbf{8.32\times10^{-2}}$ \\
 & BAGAN-GP & 204.70 & \num{2.58E-01} & \textbf{149.28} & \num{2.79E-01} & 388.60 & \num{2.54E-01} & 317.47 & \num{2.81E-01} & 529.69 & \num{5.17E-02} & 455.16 & \num{7.23E-02} \\
 & CAP-GAN & \textbf{176.47} & $\mathbf{2.59\times10^{-1}}$ & 163.90 & $\mathbf{2.83\times10^{-1}}$ & \textbf{268.31} & \num{2.70E-01} & \textbf{225.08} & $\mathbf{2.95\times10^{-1}}$ & \textbf{411.23} & $\mathbf{7.00\times10^{-2}}$ & \textbf{346.43} & \num{8.00E-02} \\ \cmidrule{2-14}

\multirow{3}{*}{100} & DCGAN & 511.31 & \num{1.83E-01} & 711.49 & \num{1.68E-01} & 703.26 & $\mathbf{2.78\times10^{-1}}$ & 811.17 & \num{2.69E-01} & 701.54 & \num{5.26E-02} & 520.11 & \num{6.32E-02} \\ 
 & BAGAN-GP & 228.19 & \num{2.54E-01} & 167.74 & \num{2.79E-01} & 416.67 & \num{2.65E-01} & 316.33 & \num{2.70E-01} & 554.70 & \num{4.72E-02} & 456.06 & \num{6.33E-02} \\
 & CAP-GAN & \textbf{168.00} & $\mathbf{2.66\times10^{-1}}$ & \textbf{160.88} & $\mathbf{2.81\times10^{-1}}$ & \textbf{286.45} & \num{2.70E-01} & \textbf{236.04} & $\mathbf{2.94\times10^{-1}}$ & \textbf{370.82} & $\mathbf{7.00\times10^{-2}}$ & \textbf{339.45} & $\mathbf{8.00\times10^{-2}}$ \\

\midrule
\multirow{3}{*}{p-value} & & \multicolumn{2}{c}{FID} & \multicolumn{2}{c}{SSIM} & \multicolumn{2}{c}{FID} & \multicolumn{2}{c}{SSIM} & \multicolumn{2}{c}{FID} & \multicolumn{2}{c}{SSIM} \\

 & CAP-GAN vs BAGAN-GP & \multicolumn{2}{c}{\num{1.3E-03}} & \multicolumn{2}{c}{\num{7.1E-01}} & \multicolumn{2}{c}{\num{5.0E-06}} & \multicolumn{2}{c}{\num{8.0E-01}} & \multicolumn{2}{c}{\num{5.0E-16}} & \multicolumn{2}{c}{\num{3.3E-02}} \\

 & CAP-GAN vs DCGAN & \multicolumn{2}{c}{\num{2.8E-12}} & \multicolumn{2}{c}{\num{2.4E-01}} & \multicolumn{2}{c}{\num{8.9E-15}} & \multicolumn{2}{c}{\num{8.1E-01}} & \multicolumn{2}{c}{\num{4.5E-11}} & \multicolumn{2}{c}{\num{1.5E-01}} \\

\bottomrule
\end{tabular}%
}
\caption{Averaged FID and SSIM for General Vision Benchmarks}
\label{tb_summary}
\end{table*}

\section{Results and Discussion}
\subsection{Datasets}
The experiments are conducted on general vision and medical imaging datasets. For general vision datasets, we consider MNIST, Fashion-MNIST, and CIFAR-10. All samples in the three datasets are resized into a uniform size, which is the same as the original sample size in CIFAR-10 (i.e. $32 \times 32$). For medical imaging datasets, we used small-scale blood cells data (Cells \cite{huang2021enhanced}) and a breast cancer cell data (BreakHis \cite{spanhol2015dataset}). We scaled down the sample size in both datasets due to the computational and time constraints. The samples in Cells and BreakHis datasets are reshaped to $64 \times 64$ and $32 \times 32$, respectively. The summary of datasets is shown in Table \ref{tab:datasets}.

\begin{table}[htb]
    \centering
    \resizebox{\linewidth}{!}{
\begin{tabular}{|c||c|c||c|c|c|c|}
\hline & &  & \multicolumn{4}{c|} { Training Samples Per Class } \\
\cline { 4 - 7 } Dataset & Resolution & Class & Min & Median & Mean & Max \\
\hline \hline MNIST & $28 \times 28$ & 10 & 6000 & 6000 & 6,000 & 6000 \\
\hline Fashion-MNIST & $28 \times 28$ & 10 & 6000 & 6000 & 6000 & 6000 \\
\hline CIFAR-10 & $32 \times 32$ & 10 & 5,000 & 5,000 & 5,000 & 5,000 \\
\hline $BreakHis^{*}$ & $700 \times 460$ & 2 & 2,480 & N/A & N/A & 5,429 \\
\hline Cells & $100 \times 101$ & 4 & 106 & 887 & 1,721 & 5,600 \\
\hline
\end{tabular}}
    
    \caption{Datasets characterstics. BreakHis only has two classes so the median and mean are N/A's.}
    \label{tab:datasets}
\end{table}

\subsection{Imbalance Rate}
The general vision datasets are balanced initially. We impose imbalance on MNIST, Fashion-MNIST, and CIFAR-10 by choosing one class as the majority class and treating the other classes as minority classes and sampling subsets for those classes. The imbalance rate is defined as the number of samples between the largest majority class and the smallest minority class. For general vision datasets, we construct imbalanced datasets using the following imbalance rates: 5, 10, 20, 50, and 100. Cells and BreakHis are originally imbalanced, we conduct the experiments on those datasets using the original imbalance rate, unless for BreakHis, we force imbalance in addition to the original imbalance rate. 

\subsection{Evaluation Metrics and Compared methods}
The Fréchet Inception Distance (FID) and Structural Similarity Index Measure (SSIM) are the metrics used for the evaluations. Lower FID or higher SSIM indicates better performance. For each class in each dataset, the model under evaluation generated 1,000 samples. Those generated samples are compared with the test samples to compute the corresponding FID and SSIM with the test set. Two baseline models are used for comparison: Conditional Deep Convolutional Generative Adversarial Networks (DCGAN)\cite{radford2015unsupervised} and the BAGAN-GP\cite{huang2021enhanced}.

\subsection{Experiment Setting}
The hyperparameter values during the training are summarized in Table \ref{tab:hyperparameter}. The compared methods are evaluated under the same experiment settings and implemented using TensorFlow 2.0 framework. We utilize NVIDIA Tesla P100 to train the models.

\begin{table}[h]
\centering

\resizebox{\linewidth}{!}{%
\begin{tabular}{lc}
\hlineB{3}
           Hyperparameter                    & Values                                                                \\ \cline{1-2}

Learning Rate (CVAE)          & 0.0006, 0.0007, 0.0008, 0.001, 0.0005                                 \\
CVAE Epoch                    & 30, 40, 50                                                            \\
Adam beta1 (CVAE)                & 0.5, 0.6, 0.7, 0.8                                                    \\ 
Learning Rate (Generator)     & 0.00005, 0.0001, 0.0002, 0.0005, 0.0008, 0.0013, 0.0015, 0.001, 0.002 \\
Learning Rate (Discriminator) & 0.00005, 0.0001, 0.0008, 0.0013, 0.0015, 0.0002, 0.002                \\
Gradient Penalty Weight                     & 5, 10                                                                 \\
Train Ratio                     & 2, 3, 4, 5, 6, 7, 8, 10                                               \\

Batch Size                    & 32, 64, 128, 256                                                      \\
Latent Dimension                    & 64, 128, 256, 512                                                     \\

\hlineB{3}
\end{tabular}%
}
\caption{Hyperparameter Optimization for CAP-GAN}
\label{tab:hyperparameter}
\end{table}

\subsection{Results for General Vision Datasets}
\label{subsec:general}

\begin{figure}[t]
\centering
\includegraphics[width=0.3\linewidth]{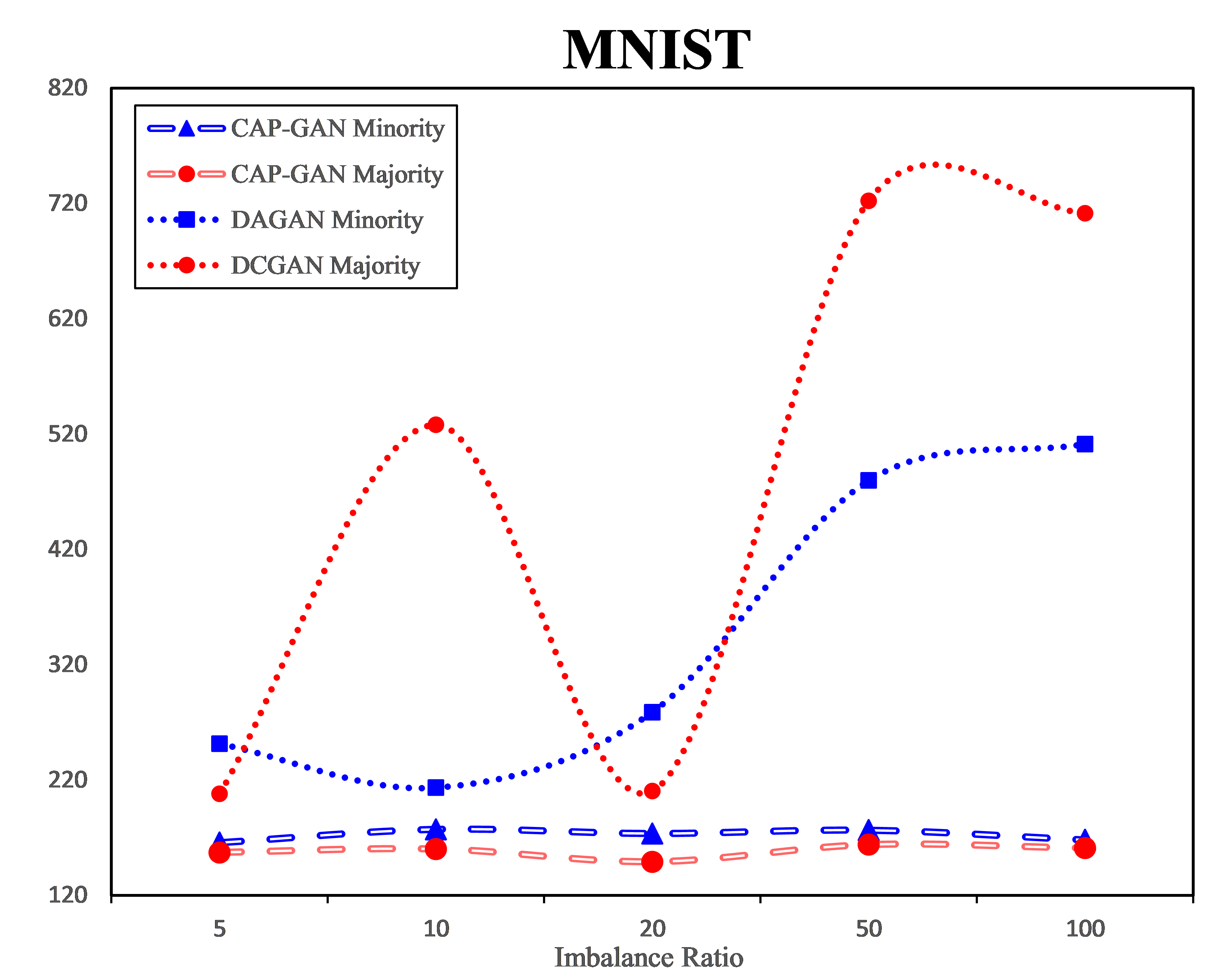}\hfill
\includegraphics[width=0.3\linewidth]{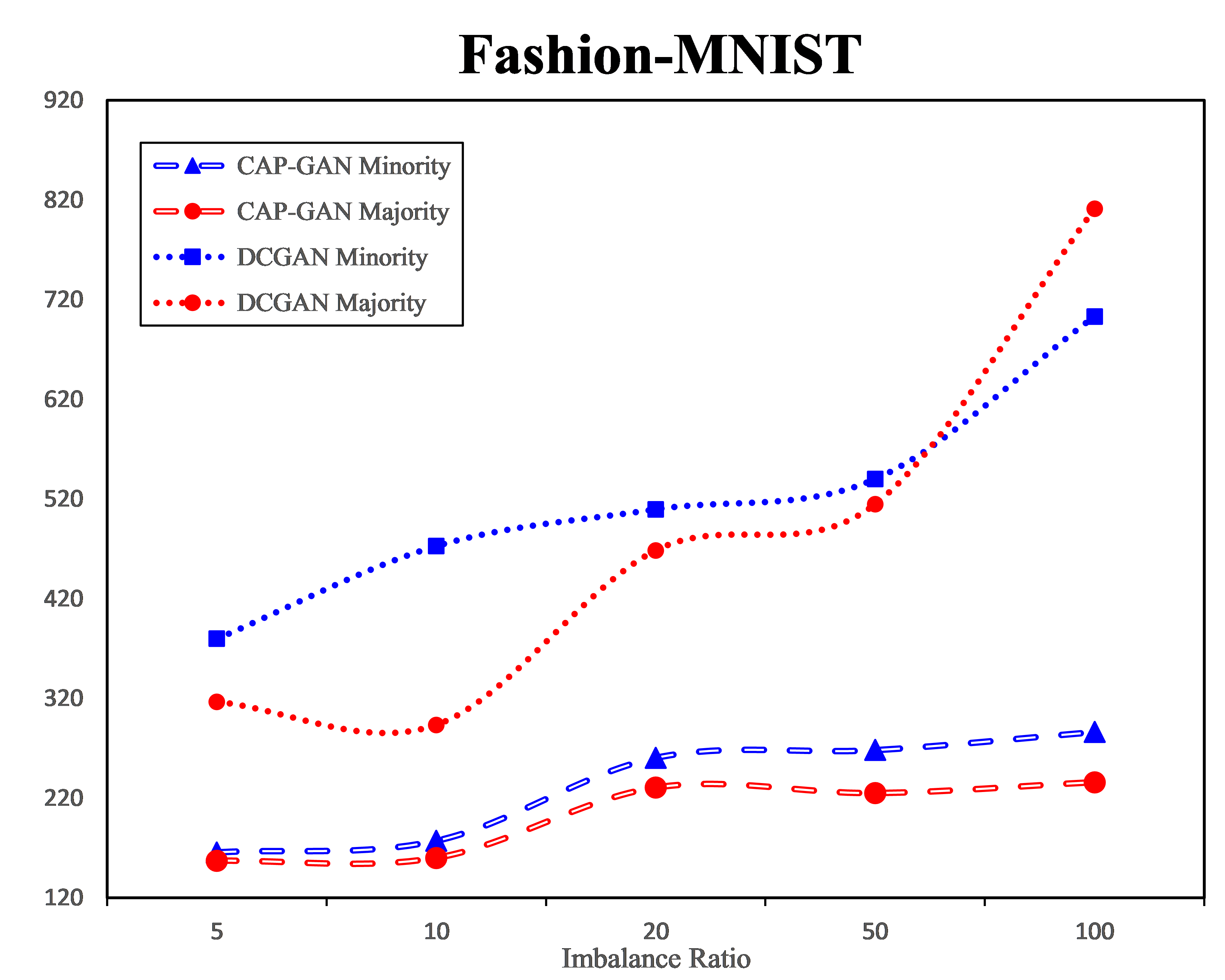}\hfill
\includegraphics[width=0.3\linewidth]{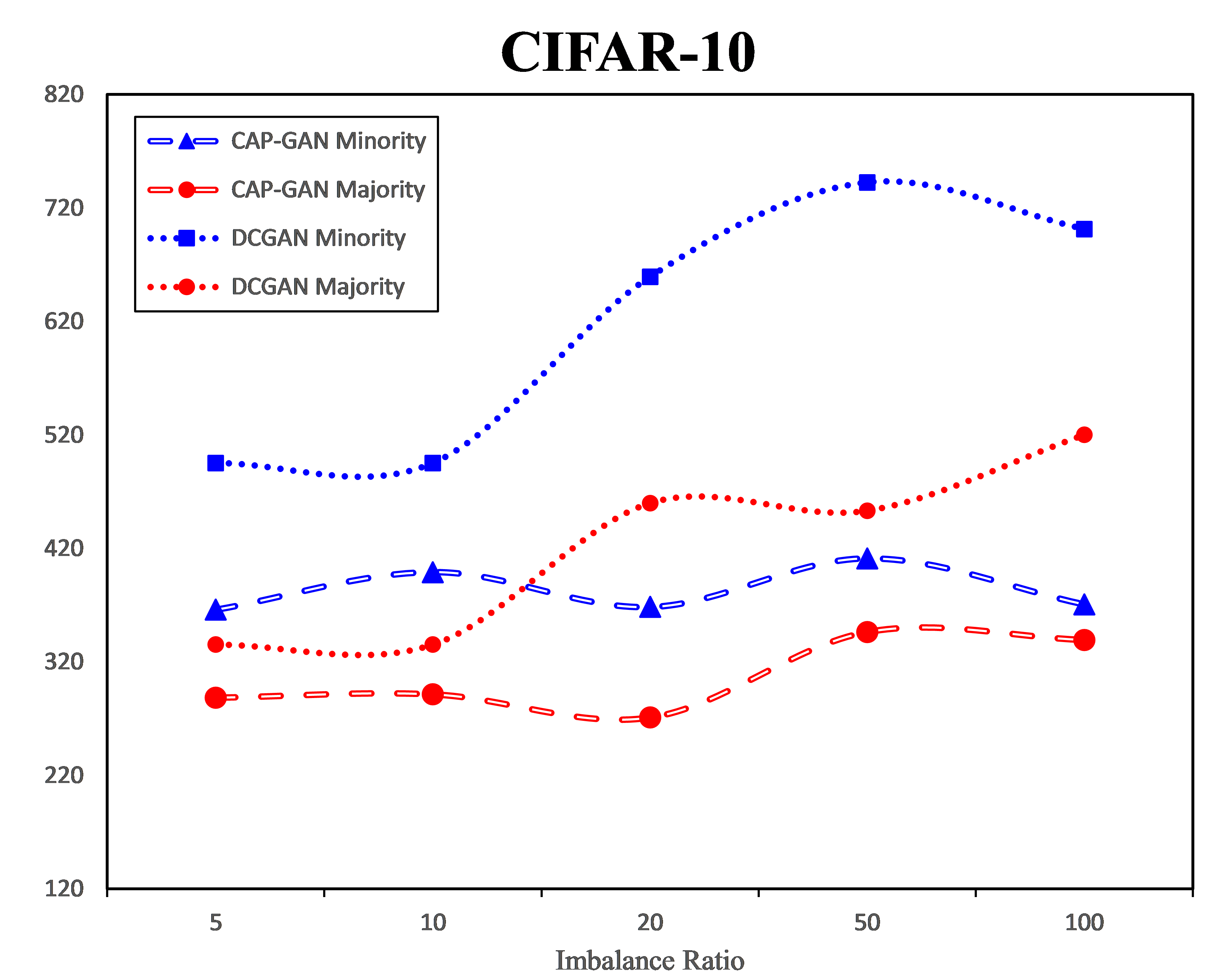}
\caption{Comparison of DCGAN and CAPGAN}
\label{fig:comp2}
\end{figure}
\begin{figure}[t]
\centering
\includegraphics[width=0.3\linewidth]{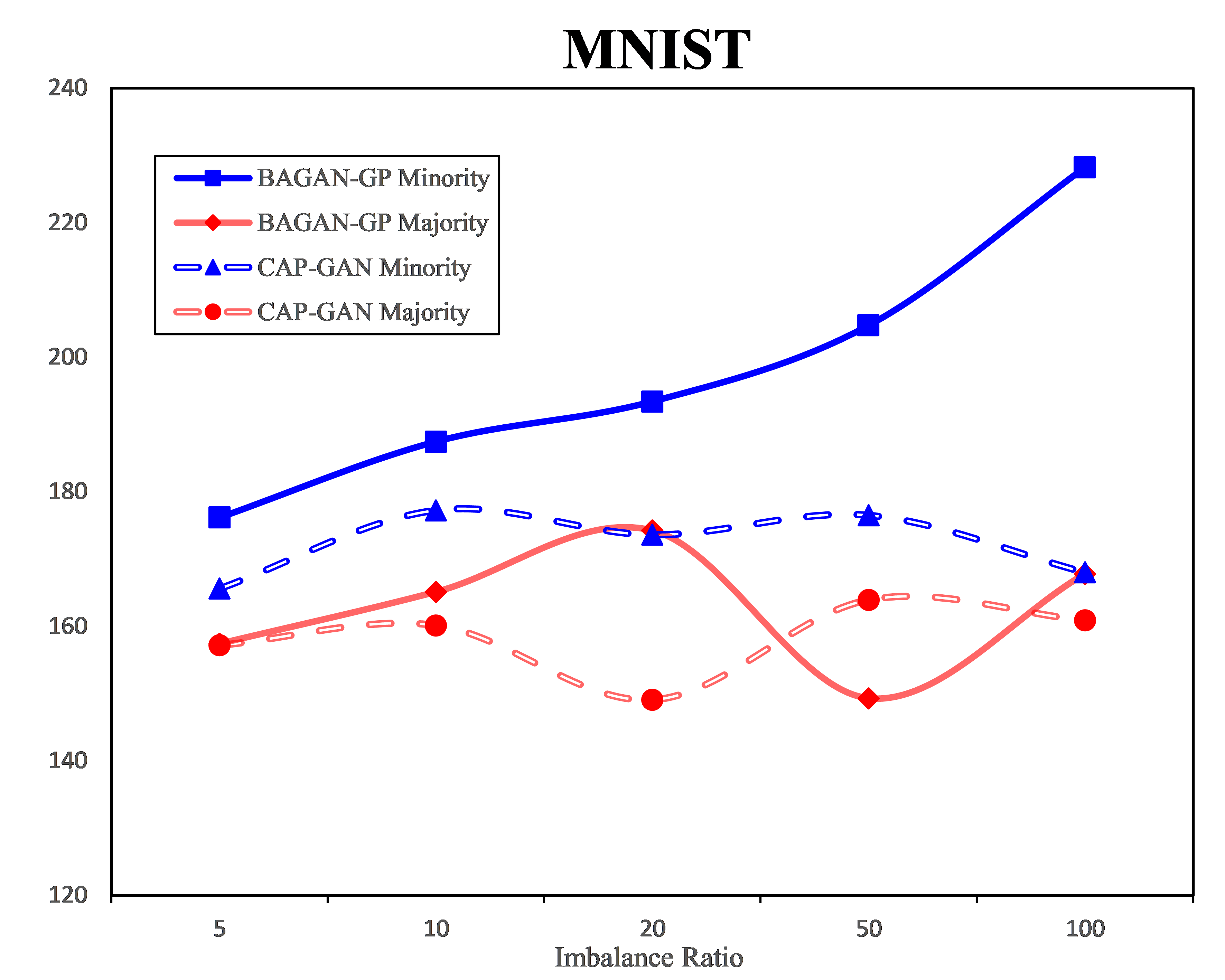}\hfill
\includegraphics[width=0.3\linewidth]{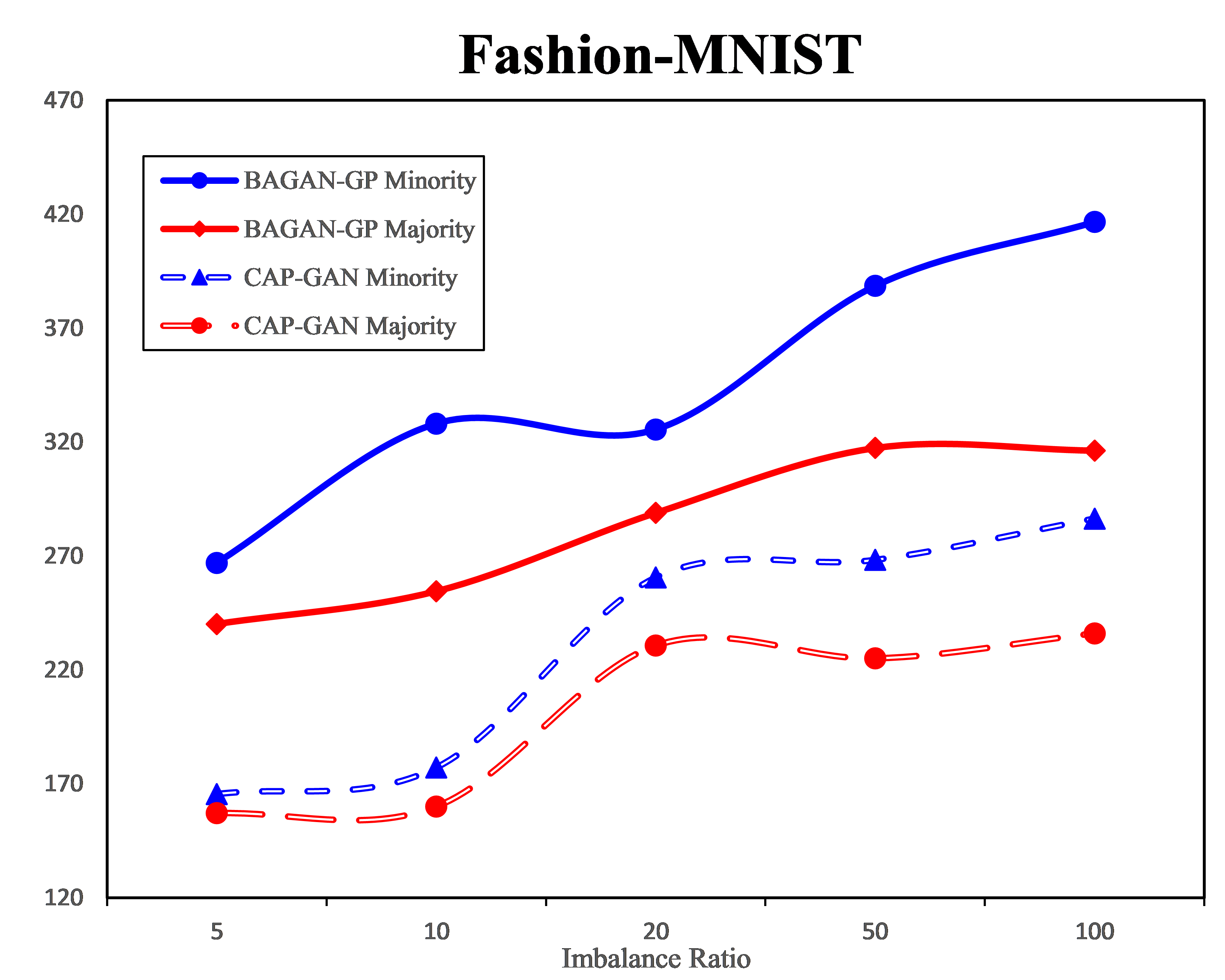}\hfill
\includegraphics[width=0.3\linewidth]{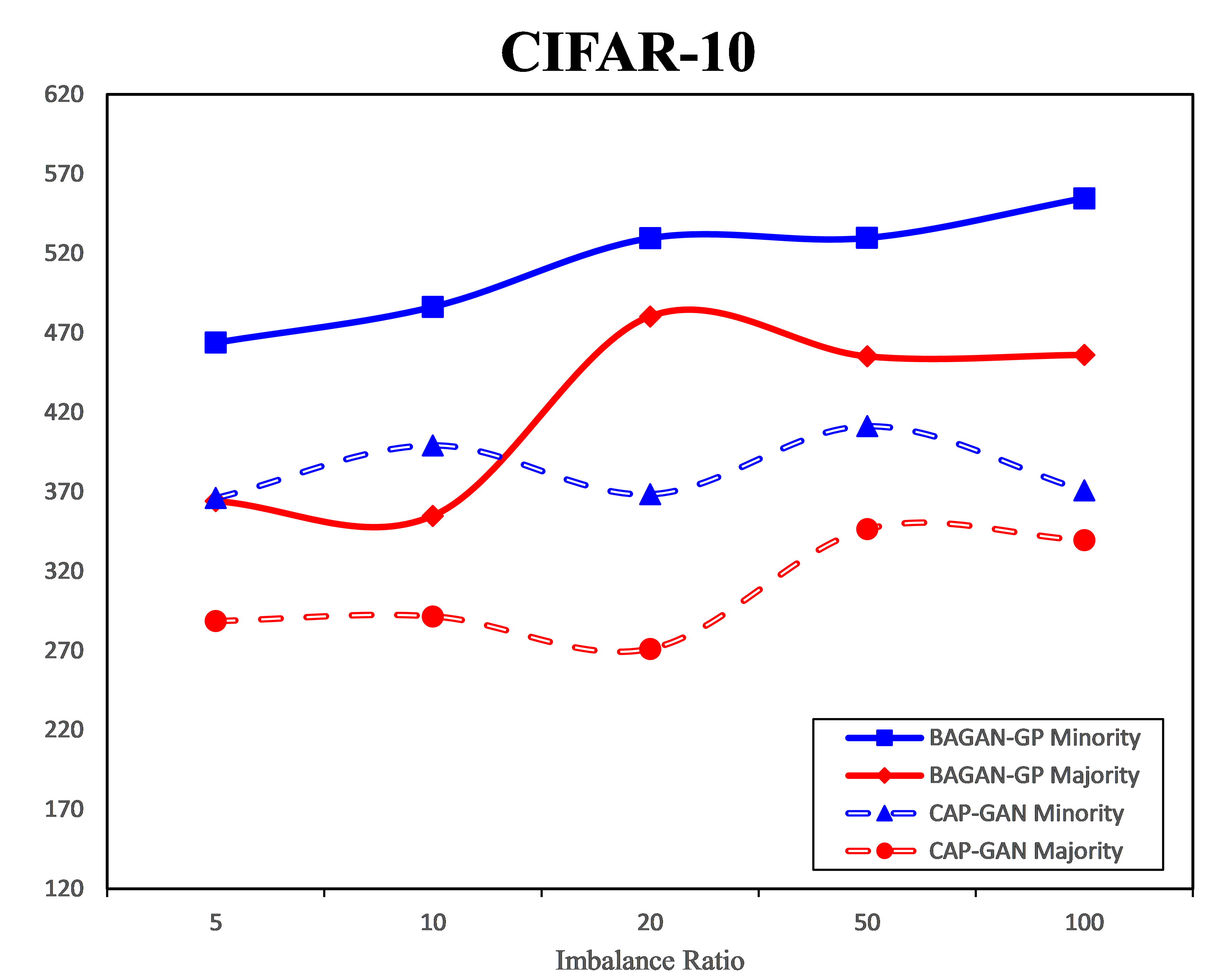}
\caption{Comparison of BAGAN-GP and CAPGAN}
\label{fig:comp1}
\end{figure}

We first conduct experiments on the low, moderate, high and extreme imbalanced versions of MNIST, Fashion-MNIST, and CIFAR-10 balanced datasets using different imbalance rates. For each dataset, there is one majority class and nine minority classes. The average values of FIDs and SSIMs for the majority and the minority class are presented in Table \ref{tb_summary}. It is clearly shown from the results that BAGAN-GP and CAPGAN outperform the DCGAN consistently across all datasets, suggesting that the countermeasures for class imbalance are effective. It also indicates that generative models which are proved to be successful on balanced datasets could not handle class imbalance. We further observe that CAPGAN achieves superior results than BAGAN-GP in almost all experiment settings, which indicates that CAPGAN is much more powerful than BAGAN-GP as a generative model for imbalanced data. We also note, as shown in Figures \ref{fig:comp2} and \ref{fig:comp1}, that the compared methods produce unstable results when the imbalance rate increase. In particular, the FID of BAGAN-GP and DCGAN increase significantly as the imbalance rate becomes higher in all three datasets. In contrast, the FID of CAPGAN remains at a steady level or increases much lower as the imbalance rate increases. Therefore, CAPGAN is proven as a powerful generative model that is able to maintain a low FID in the presence of high and extreme imbalance rates (i.e., 50 and 100). As for SSIM, although the improvements are not as evident as the FID, CAPGAN managed to score higher SSIM scores under most experiment configurations than DCGAN and BAGAN-GP. In terms of SSIM evaluation metric, although the improvements are not as evident as the FID, CAPGAN is able to attain higher scores under most experiment configurations compared to the state-of-the-arts. We believe the CVAE initialization and the proposed pre-training help CAPGAN to achieve such results. Furthermore, using the Student's t-test, we investigate whether the results that are produced by CAPGAN are significantly different to the state-of-the-art. The statistical results show that the p-value in all tests for FID is less than 0.05, which rejects the null hypothesis that CAPGAN and the compared methods have equal performance.


\begin{figure}[t]
     \centering
     \begin{subfigure}[t]{0.24\linewidth}
         \centering
         \includegraphics[width=\linewidth]{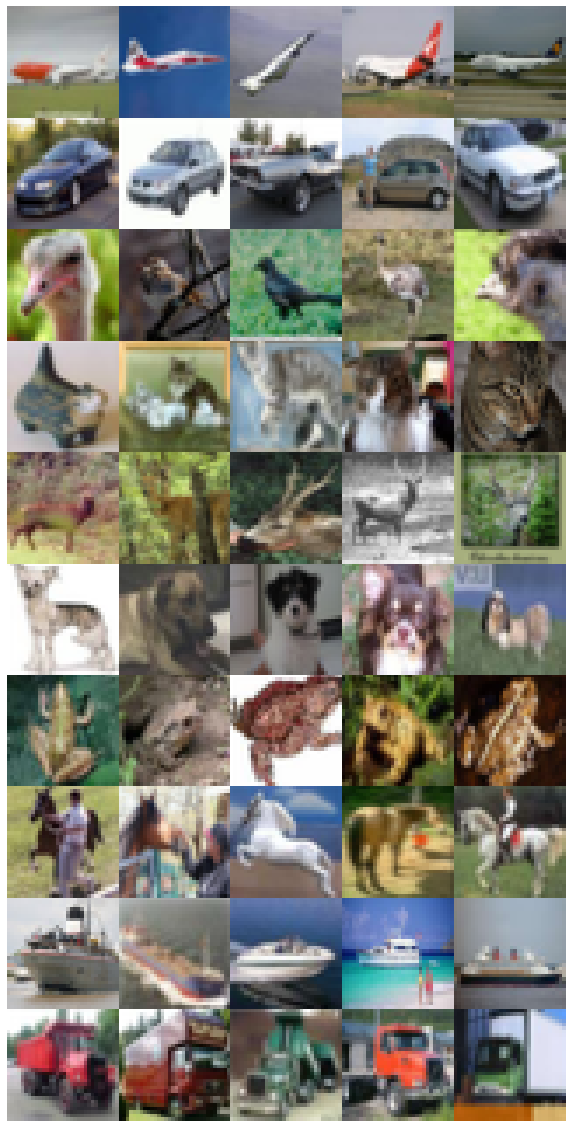}
         \caption{Original Image}
         \label{fig:CIFAR-ORIGINAL}
     \end{subfigure}
     \hspace{-5mm} 
     \hfill
     \begin{subfigure}[t]{0.24\linewidth}
         \centering
         \includegraphics[width=\linewidth]{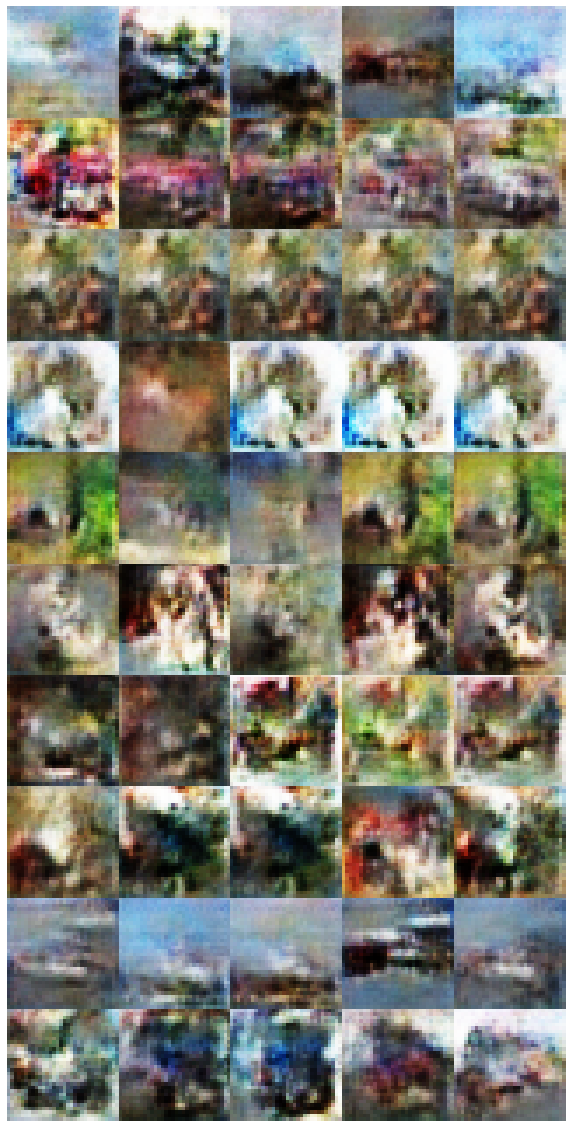}
         \caption{DCGAN}
         \label{fig:CIFAR-DCGAN}
     \end{subfigure}
     \hspace{-5mm}
     \hfill
     \begin{subfigure}[t]{0.24\linewidth}
         \centering
         \includegraphics[width=\linewidth]{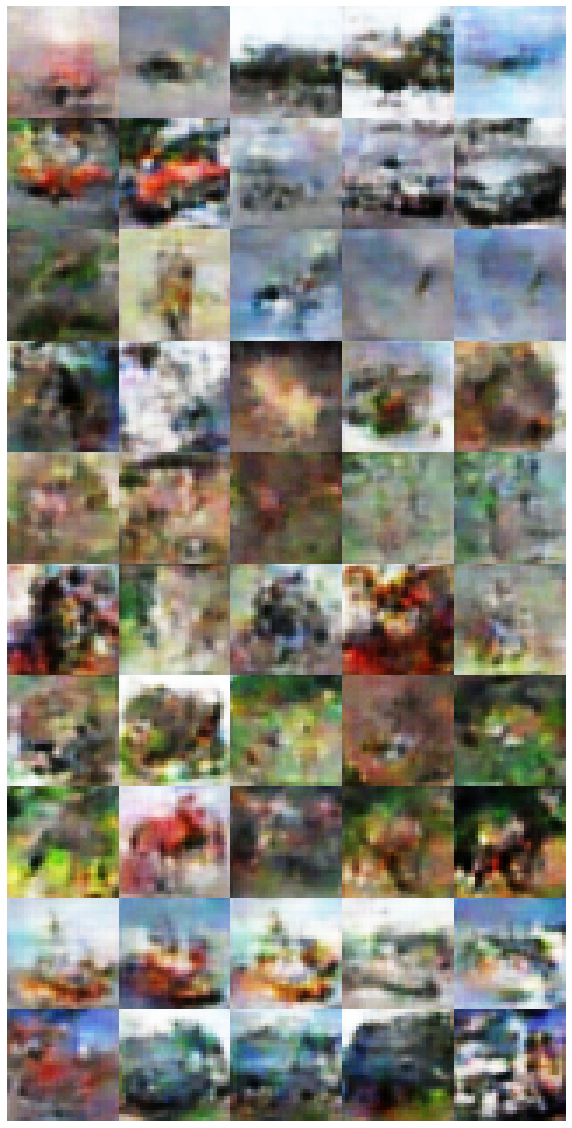}
         \caption{BAGAN-GP}
         \label{fig:CIFAR10-BAGAN-GP}
     \end{subfigure}
     \hspace{-5mm}
     \hfill
     \begin{subfigure}[t]{0.24\linewidth}
         \centering
         \includegraphics[width=\linewidth]{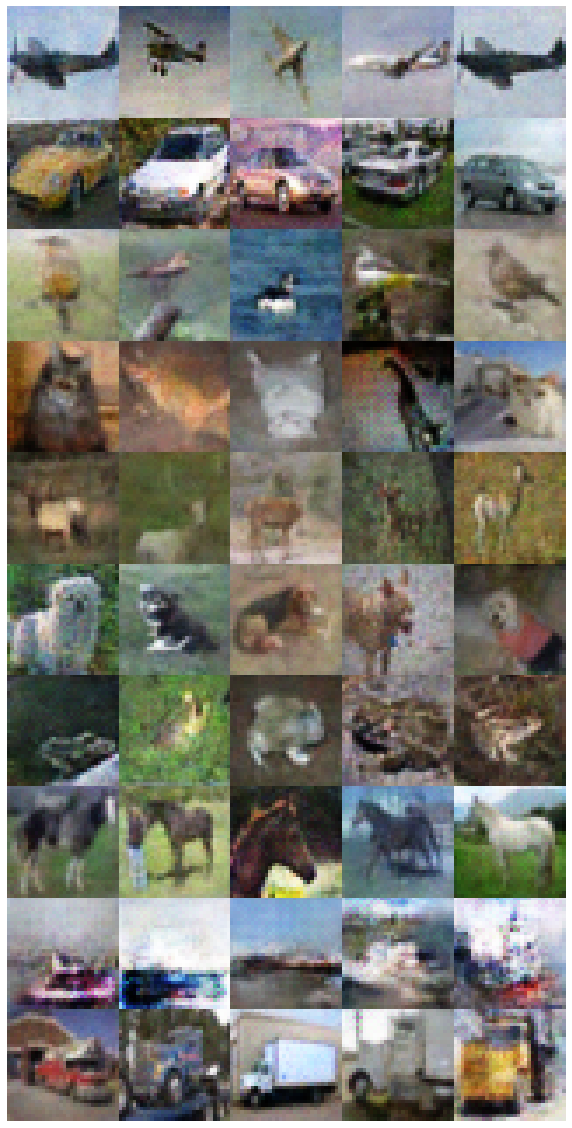}
         \caption{CAPGAN}
         \label{fig:CIFAR10-CAPGAN}
     \end{subfigure}
     \hspace{-5mm}
        \caption{Generated Images for CIFAR-10 with an Imbalanced Rate of 100}
        \label{fig:CIFAR10-Comparsion of generated images}
\end{figure}
We investigate the generated images from DCGAN, BAGAN-GP, and CAPGAN, along with the original images on CIFAR-10. As shown in Figure \ref{fig:CIFAR10-Comparsion of generated images}, the generated images are all from the minority classes. Both DCGAN and BAGAN-GP produce very blurring images when the imbalance rate is extreme (i.e. 100). Most of the generated samples using these two models lose details and textures, and several samples look like unrecognizable objects (i.e., noises). Further, they are lack of diversity, which indicates that DCGAN and BAGAN-GP suffers from mode collapse. On the contrary, CAPGAN generates more realistic samples that capture the details and the textures of the objects. Each sample is easily recognized regarding its original class. Furthermore, CAPGAN produces samples of high diversity, which is crucial for reducing the likelihood of overfitting when using the CAPGAN for oversampling and achieving a better performance at high and extreme imbalance rates.


\subsection{Results for Medical Imaging Benchmarks}
\label{subsec:medicalres}

\begin{table}
\centering
\resizebox{\columnwidth}{!}{%
\begin{tabular}{rlrrrr}
\toprule
                  &                 & \multicolumn{4}{c}{\textbf{Cells}}                \\ \cmidrule{2-6} 
                  &                 & \multicolumn{2}{c}{avg(Minority)} & \multicolumn{2}{c}{Majority} \\
                  &                 & FID            & SSIM        & FID            & SSIM        \\
Imbalance Rate & Model &                &             &                &             \\ \midrule
\multirow{3}{*}{52.83} & DCGAN            & 438.48         &  \num{3.75E-01}       & 266.99         & \num{4.46E-01}        \\
 & BAGAN-GP            & 260.79        & \num{3.51E-01}        & 247.34         & \num{4.27E-01}        \\
 & CAP-GAN            & \textbf{228.76}         & \num{3.48E-01}        & \textbf{160.00}         & \num{4.35E-01}        \\ 
\midrule

\multirow{2}{*}{Improvement on FID (\%)} & CAP-GAN vs BAGAN-GP & \multicolumn{2}{c}{$12.28\uparrow$} & \multicolumn{2}{c}{$35.31\uparrow$} \\
 & CAP-GAN vs DCGAN & \multicolumn{2}{c}{$47.83\uparrow$} &	\multicolumn{2}{c}{$40.07\uparrow$} \\
\bottomrule
\end{tabular}
}
\caption{Averaged FID and SSIM for Cells}
\label{tb_cells}
\end{table}
\begin{table}
\centering

\resizebox{\columnwidth}{!}{%
\begin{tabular}{rlrrrr}
\toprule
                  &                 & \multicolumn{4}{c}{\textbf{BreakHis}}                                \\ \cmidrule{2-6} 
                  &                 & \multicolumn{2}{c}{Minority} & \multicolumn{2}{c}{Majority} \\
                  &                 & FID            & SSIM        & FID            & SSIM        \\
Imbalance Rate & Model & & & &             \\ \midrule

\multirow{3}{*}{2.19}
    & DCGAN & 456.39         & \num{5.66E-02}  & 294.23         & \num{6.34E-02}    
\\
& BAGAN-GP &  251.00         & \num{6.76E-02}        & 231.76         & \num{6.47E-02}        \\
& CAP-GAN & \textbf{237.47}         & \num{5.91E-02}        & \textbf{216.01}         & \num{6.06E-02}        \\ \cmidrule{2-6}

\multirow{3}{*}{10}
& DCGAN  & 565.67        &\num{5.25E-02}  & 416.73         & \num{5.86E-02}       \\
& BAGAN-GP & 211.28         & \num{8.95E-02}       &210.61         &\num{9.30E-02}       \\
& CAP-GAN & \textbf{203.60}        & \num{5.94E-02}        & \textbf{181.82}         & \num{5.95E-02}       \\

 \midrule
\multirow{2}{*}{Improvement on FID (\%) of 2.19} & CAP-GAN vs BAGAN-GP & \multicolumn{2}{c}{$5.39\uparrow$} &	\multicolumn{2}{c}{$6.79\uparrow$} \\
 & CAP-GAN vs DCGAN & \multicolumn{2}{c}{$47.97\uparrow$} &	\multicolumn{2}{c}{$26.58\uparrow$} \\ \cmidrule{2-6}
 
\multirow{2}{*}{Improvement on FID (\%) of 10} & CAP-GAN vs BAGAN-GP & \multicolumn{2}{c}{$3.63\uparrow$} &	\multicolumn{2}{c}{$13.67\uparrow$} \\
 & CAP-GAN vs DCGAN & \multicolumn{2}{c}{$64.01\uparrow$} &	\multicolumn{2}{c}{$56.37\uparrow$} \\
\bottomrule
\end{tabular}
}
\caption{FID and SSIM for BreakHis}
\label{tb_breakhis}
\end{table}

Medical imaging data are often imbalanced due to the high cost of generating real images. The generative models play an important role to produce synthetic images at medical applications. The results for medical imaging data are presented in Table \ref{tb_cells} and Table \ref{tb_breakhis}. Cells is a small-scaled medical dataset with a high imbalance rate (i.e. 52.83). Similar to the results of general vision datasets. Firstly, we investigate the impact of the class imbalance on the baseline DCGAN. We find that BAGAN-GP and CAPGAN achieve better performance than DCGAN on almost all metrics, especially FID. It shows that the powerful generative architecture on traditional and balanced datasets could not adapt to more challenging and high imbalanced medical imaging data, especially for the minority classes, which is considered as the class of interest. Secondly, we test the performance of the proposed CAPGAN against the-state-of-arts BAGAN-GP. CAPGAN outperforms the BAGAN-GP in FID. The improvements on FID are around 12.78\% and 35.31\% for minority and majority classes, respectively. The boost on SSIM is not as significant as FID, where CAPGAN and BAGAN-GP achieved comparable results.

For BreakHis dataset, although the original imbalance rate is lower than Cells, it is challenging in other aspects because it contains images of different scales, such as the objects in different images might be collected at a different magnification rate. As presented in Table \ref{tb_breakhis}, BAGAN-GP and CAPGAN consistently outperform the DCGAN, especially under a high imbalance rate. The results suggest the effectiveness of class imbalance countermeasures. Furthermore, CAPGAN achieves better FID than BAGAN-GP under both imbalance rates. As for SSIM, all three models obtained very low SSIM, so it would be pointless to analyze the statistics regarding the SSIM. We believe that the variety of scales in the datasets lead to unsatisfying performance for all three models since they do not have any technique to deal with samples of different scales. 



\begin{figure}[t]
     \centering
     \begin{subfigure}[t]{0.24\linewidth}
         \centering
         \includegraphics[width=\linewidth]{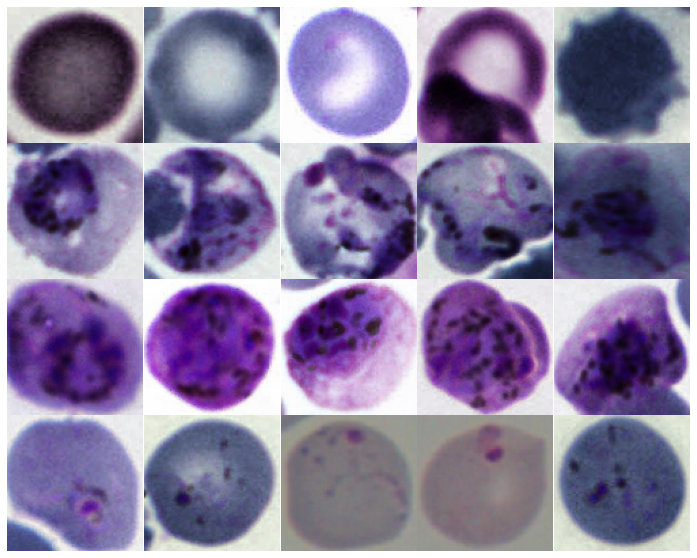}
         \caption{Original Image}
         \label{fig:CELLS-ORIGINAL}
     \end{subfigure}
     \hspace{-5mm} 
     \hfill
     \begin{subfigure}[t]{0.24\linewidth}
         \centering
         \includegraphics[width=\linewidth]{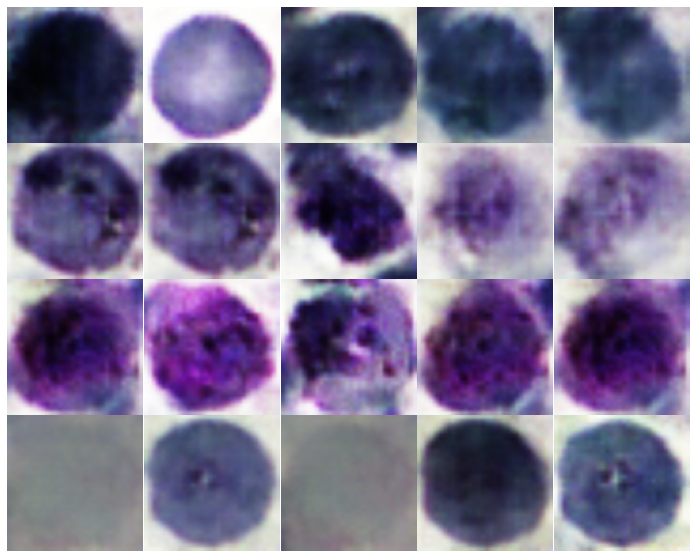}
         \caption{DCGAN}
         \label{fig:CELLS-DCGAN}
     \end{subfigure}
     \hspace{-5mm}
     \hfill
     \begin{subfigure}[t]{0.24\linewidth}
         \centering
         \includegraphics[width=\linewidth]{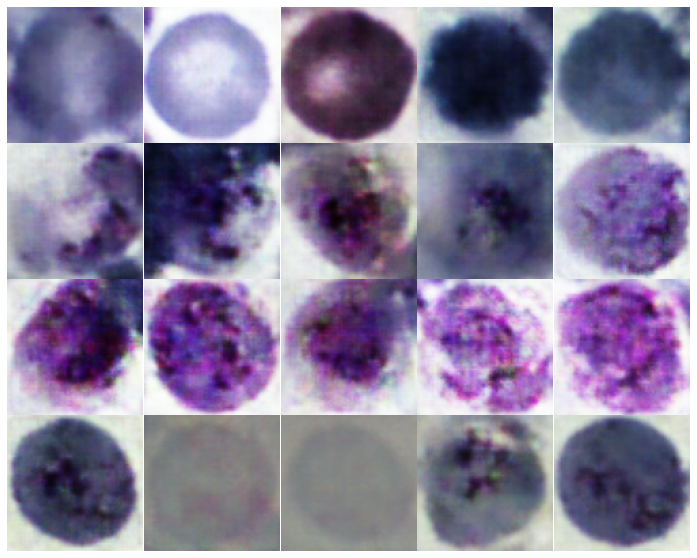}
         \caption{BAGAN-GP}
         \label{fig:CELLS-BAGAN-GP}
     \end{subfigure}
     \hspace{-5mm}
     \hfill
     \begin{subfigure}[t]{0.24\linewidth}
         \centering
         \includegraphics[width=\linewidth]{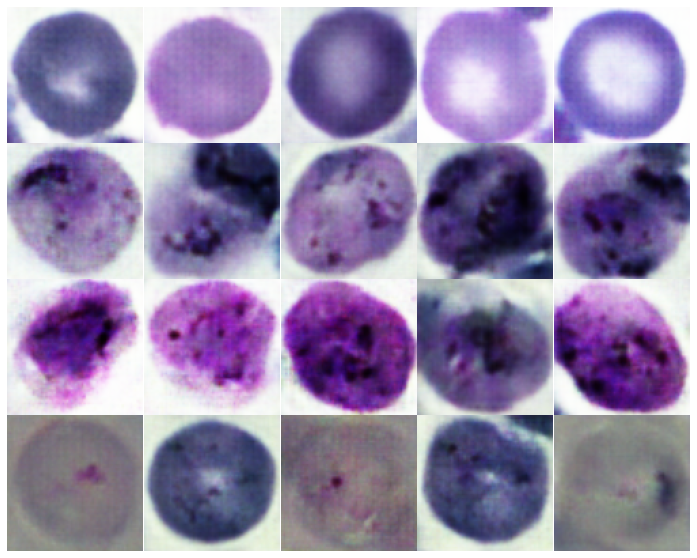}
         \caption{CAPGAN}
         \label{fig:CELLS-CAPGAN}
     \end{subfigure}
     \hspace{-5mm}
        \caption{Generated Images for Cells}
        \label{fig:CELLS-Comparsion of generated images}
\end{figure}

We illustrate the generated images from DCGAN, BAGAN-GP, and CAPGAN in the Cells. As shown in Figure \ref{fig:CELLS-Comparsion of generated images}, the generated samples from DCGAN and BAGAN-GP are poor in their quality due to a lack of textures and details. Besides, we can observe many irregular dark or grey shapes on those generated images, which cover a large portion of the cell and make the cell unrecognizable. However, the generated images by CAPGAN are more realistic and diverse cell images. The details and textures are clear with small noises covering cell's body. Furthermore, the cell images have richer and more diverse colours than those generated by DCGAN and BAGAN-GP, which are dark and have low contrast. 


\section{Conclusion}
In this work, we propose a method to mitigate the problem of class imbalance in the imaging domain by utilizing generative models. The proposed method CAPGAN facilitates a conditional convolutional variational autoencoder (CVAE), which has an embedding component to perform training in a supervised fashion. A new objective function is applied to the CVAE training to improve the generative and reconstruction ability. Moreover, we present several pre-training strategies which could lead to produce balanced weights for the generative model. The generator and discriminator in CAPGAN are initialized by the pre-trained components in the CVAE and are trained in an adversarial setting. A gradient penalty term is added to the loss function of the discriminator to help stabilize the GAN training. We demonstrate the efficiency of CAPGAN on various datasets, including hand-crafted imbalanced datasets from general vision datasets and two imbalanced medical imaging datasets. We compare our proposed model with DCGAN and BAGAN-GP. The results show that CAPGAN outperforms these two methods by generating higher quality images given imbalanced datasets. Empirical results indicate that CAPGAN can retain high performance as the imbalance rate increases and can deliver acceptable results even under extreme imbalanced situations. 

\clearpage

\bibliographystyle{named}
\bibliography{ijcai22-multiauthor}


\end{document}